# Intelligent bear deterrence system based on computer vision: Reducing human–bear conflicts in remote areas

**Pengyu Chen[1,2], Teng Fei[1,5], John A. Kupfer[2], Yunyan Du[3], Jiawei Yi[3], and Yi Li[4]**

[1]*School of Resources and Environmental Sciences, Wuhan University, Wuhan 430070, China*
[2]*Department of Geography, University of South Carolina, Columbia, SC 29208, USA*
[3]*State Key Laboratory of Resources and Environmental Information System, Beijing 100101, China*
[4]*Institute of Zoology, Chinese Academy of Sciences, Beijing 100101, China*

***Abstract:*** Human–bear conflicts on the Tibetan Plateau threaten both local livelihoods and the conservation of Tibetan brown bears (*Ursus arctos pruinosus*). To address this challenge, we developed a low-power, network-independent deterrence system that combines computer vision with Internet of Things (IoT) hardware. The system integrates a YOLOv5-MobileNet detection model deployed on a low-power edge artificial intelligence (AI) board with a solar-powered bear spray device. We compiled a data set of 1,243 wildlife images (including 795 bears with 100 infrared captures for nighttime detection, plus other common objects and animals such as mastiffs, yaks, humans, and vehicles), from which 80% were used for training and 20% for validation. Validation showed robust performance (mean average precision = 91.4%, recall = 93.6%). In 100 controlled activation tests involving simulated approaches by bears, humans, and other animals, the spray deployed within 0.2 seconds of detection with 97.2% accuracy, confirming timely and reliable responses. A 30-day field trial in Zadoi County, Qinghai Province, China, recorded 3 successful deterrence events without false activations. By using energy-efficient components and ensuring continuous and stable system operation, this solution provides a practical, sustainable, and scalable approach to mitigating human–bear conflicts, effectively enhancing human safety and bear conservation in remote areas without network or grid coverage.

***Key words:*** bear detection and deterrence, brown bear, computer vision, deterrence systems, human–bear conflict, Internet of Things, Tibetan Plateau, *Ursus arctos pruinosus*

Conflicts between humans and Tibetan brown bears (*Ursus arctos pruinosus*) on the Tibetan Plateau are increasing, threatening both the conservation of this protected species and the safety and livelihoods of local herding communities (Dai et al. 2022). Encounters often result in livestock losses, property damage, and occasional human injuries or fatalities, especially during seasonal migrations when settlements are left vulnerable (Dai et al. 2020a); this intensifies tensions between conservation efforts and community safety (Dai et al. 2019, 2020b).

Meanwhile, many rural regions in the Tibetan Plateau are lacking internet access and reliable electricity because of topographic constraints and sparse population density (Tang et al. 2023), making cloud computing and mobile phone notifications impractical. Existing mitigation solutions such as mesh wire fences, domestic guard dogs, lighting firecrackers, or even hunting offer only temporary relief and are fraught with practical, ecological, and legal challenges, underscoring the need for innovative, sustainable approaches (Xufang et al. 2018; Dai et al. 2020a, 2022).

To address these challenges comprehensively, we leveraged recent advancements in Computer Vision (CV) and Internet of Things (IoT) technologies to develop a sustainable and effective bear deterrent method that offers a scalable and sustainable solution for enhancing safety and conservation in remote communities. Specifically, we sought to design a system that (1) utilizes a robust, offline bear detection mechanism, (2) employs an effective bear deterrent, and (3) is solar-powered to ensure continuous performance in remote electricity-scarce regions.

Artificial intelligence (AI) and CV have offered new methods to detect and survey animal populations with enhanced accuracy in wildlife monitoring. Early studies demonstrated the feasibility of using deep learning for wildlife detection, such as when monitoring polar bears (*U. maritimus*) in challenging conditions (Sorensen et al. 2017), and subsequent work has refined

[5]email: feiteng@whu.edu.cn

these methods, for instance, by tailoring the "You Only Look Once" (YOLO) architecture for specialized wildlife detection tasks (Roy et al. 2023). Although these advances are significant, many systems for animal monitoring have not focused on power efficiency nor the development of integrated deterrents, which are critical for practical conflict mitigation in remote areas.

A growing body of research highlights the shift toward real-time monitoring systems with automated decision-making capabilities, particularly through use of IoT and edge AI. These systems process data directly on a device in the field, overcoming the limitations of network availability. For example, studies have demonstrated the feasibility of running deep learning models for image classification on low-power edge devices such as the Raspberry Pi (Velasco-Montero et al. 2018) and have developed customized camera-trap platforms that perform object classification on the edge (Zualkernan et al. 2022, Nazir and Kaleem 2024). Such real-time systems have already been deployed to mitigate human–wildlife conflict with other large carnivores, such as using a camera-based alert system to monitor endangered tigers (*Panthera tigris*; Dertien et al. 2023). In remote areas, these alerts can even be transmitted via satellite networks (Whytock et al. 2021). Beyond simple alerts, some systems have been developed to take automated action; one example is the Felixer grooming trap, which uses sensors to deliver a targeted dose of lethal toxin to control invasive predators, such feral cats (*Felis catus*) in Australia (Moseby et al. 2020).

These developments highlight the potential of CV and IoT-based systems for advancing wildlife monitoring and management. However, most current systems emphasize detection rather than active deterrence, and they typically assume stable electricity and network access, conditions that are rarely available on the Tibetan Plateau. To address these gaps, we developed and tested an intelligent bear deterrence system that integrates lightweight computer vision models with low-power IoT hardware and a solar-powered bear spray device. Our objectives were to (1) design an embedded detection system capable of identifying bears in real time without network connectivity, (2) evaluate the reliability and safety of an automated spray module, and (3) test system performance through controlled trials and field deployment in conflict-prone communities of Qinghai Province. This study represents an AI-driven, nonlethal deterrence system aimed at reducing human–bear conflict in remote, off-grid environments.

## Methods

Our method was developed in 2 phases that involved (1) training a bear detection model on the MaixHub platform, which is optimized for the K210 board (Sipeed, Shenzhen, Guangdong, China) and other IoT devices (Sipeed 2025), and (2) designing a detection and deterrence system by embedding the trained model into hardware that can be installed in the field (Fig. 1). These steps are described in greater detail below.

### *Bear detection model*

To develop the bear detection model, we assembled a data set of 1,243 wildlife images representing species commonly encountered on the Tibetan Plateau. Images were collected from field surveys and online sources. Although the system is intended primarily for use during herders' seasonal relocations when human presence is minimal, we included human images to ensure the model could distinguish people and thereby reduce the risk of accidental spraying. The data set comprised 795 bear images, 160 mastiffs (*Canis familiaris*), 150 yaks (*Bos mutus*), 145 humans, 140 cats, 125 motorcycles, 100 cars, and 80 antelopes (*Pantholops hodgsonii*). To improve nocturnal performance, >100 bear images were infrared captures. Figure 2 summarizes the data set-by-image counts across categories.

We then annotated images with bounding boxes and fine-tuned them on the MaixHub platform using the YOLOv5 (You Only Look Once, version 5; https://docs.ultralytics.com/models/yolov5, Ultralytics, Frederick, Maryland, USA) model with MobileNet (https://github.com/tensorflow/models/blob/master/research/slim/nets/mobilenet/README.md, Google LLC, Mountain View California, USA) as its backbone to ensure efficient and accurate bear detection on low-power hardware (Howard et al. 2017, Jocher et al. 2022). YOLOv5 is a well-established object detection framework known for balancing processing speed and detection accuracy, making it particularly suitable for real-time embedded applications. MobileNet's compact architecture substantially reduces computation while maintaining reliable performance, meeting the memory and processing constraints of the K210 board.

The workflow begins by capturing input images, which are processed by the YOLOv5 model embedded in the K210 board. The model analyzes the input, identifies bears in real time, and generates bounding boxes to pinpoint their locations (Fig. 3). The key training parameters used in the fine-tuning process are summarized in Table 1.

### *Deterrence system implementation*

The bear deterrence system consists of 2 primary components: (1) a core sensing platform and IoT-based sensor, and (2) a specialized bear deterrent spray device. The core sensing platform utilizes the Kendryte K210 chip on the Sipeed MaixDuino board, which was selected for its integrated neural processing unit (NPU) and low power consumption (see Table 2 for key hardware specifications). The system used a GC0328 complimentary metal-oxide-semiconductor (CMOS) camera module connected via a 24-pin digital video port (DVP) interface to provide a continuous video stream for real-time detection. The camera offered a 90° horizontal and 60° vertical field of view, providing broad coverage around herders' homes for early bear detection. The module operated at approximately 30 frames per second (fps), with sensitivity settings adjusted to balance image clarity and energy efficiency in remote, off-grid environments. The K210 runs the YOLOv5-based bear detection model described above, enabling on-board image inference without relying on cloud services. A glossary of key technical terms is included in Table 3.

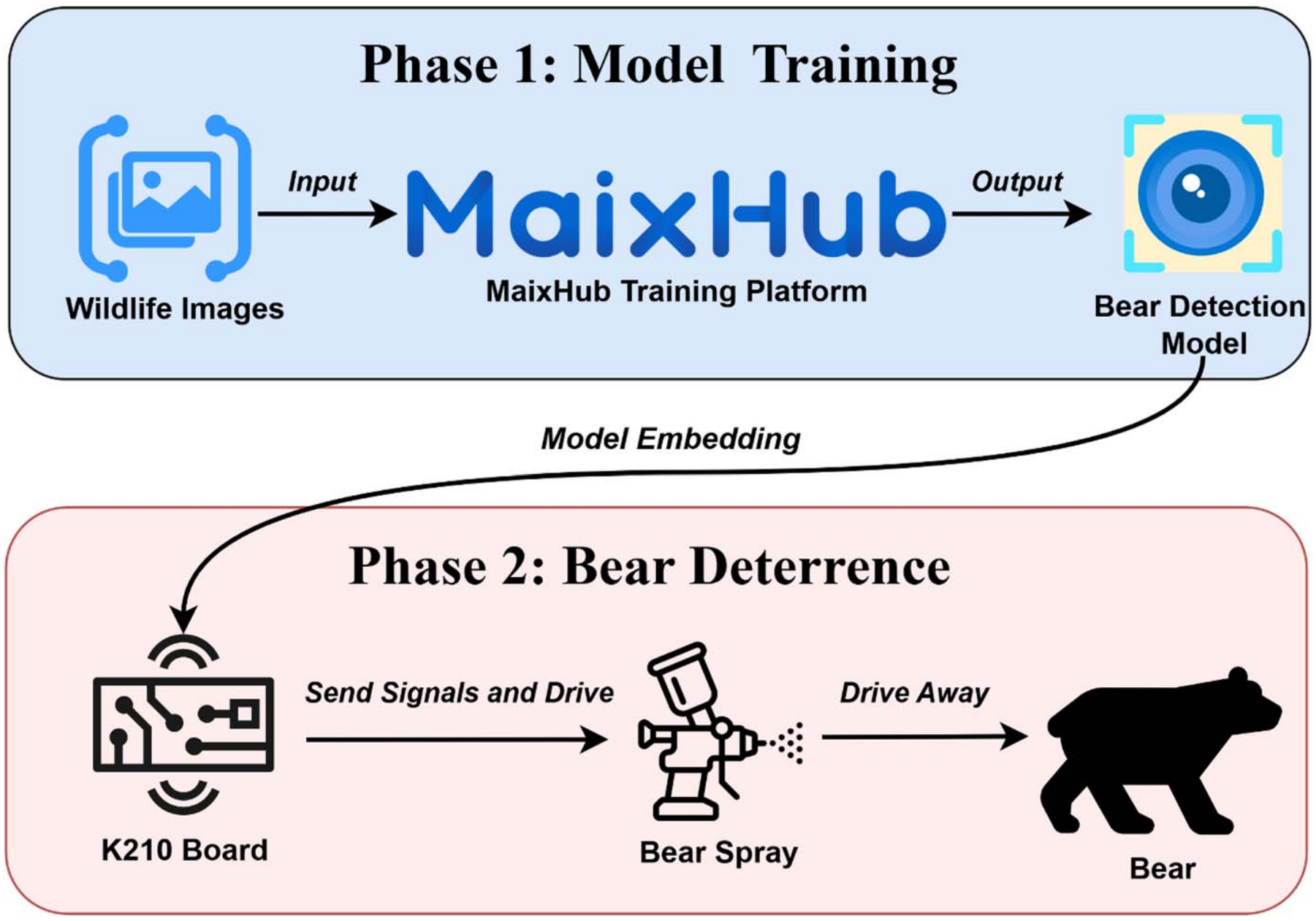

**Fig. 1. Two-phase workflow diagram for an intelligent bear deterrence system to reduce human–bear conflicts.**

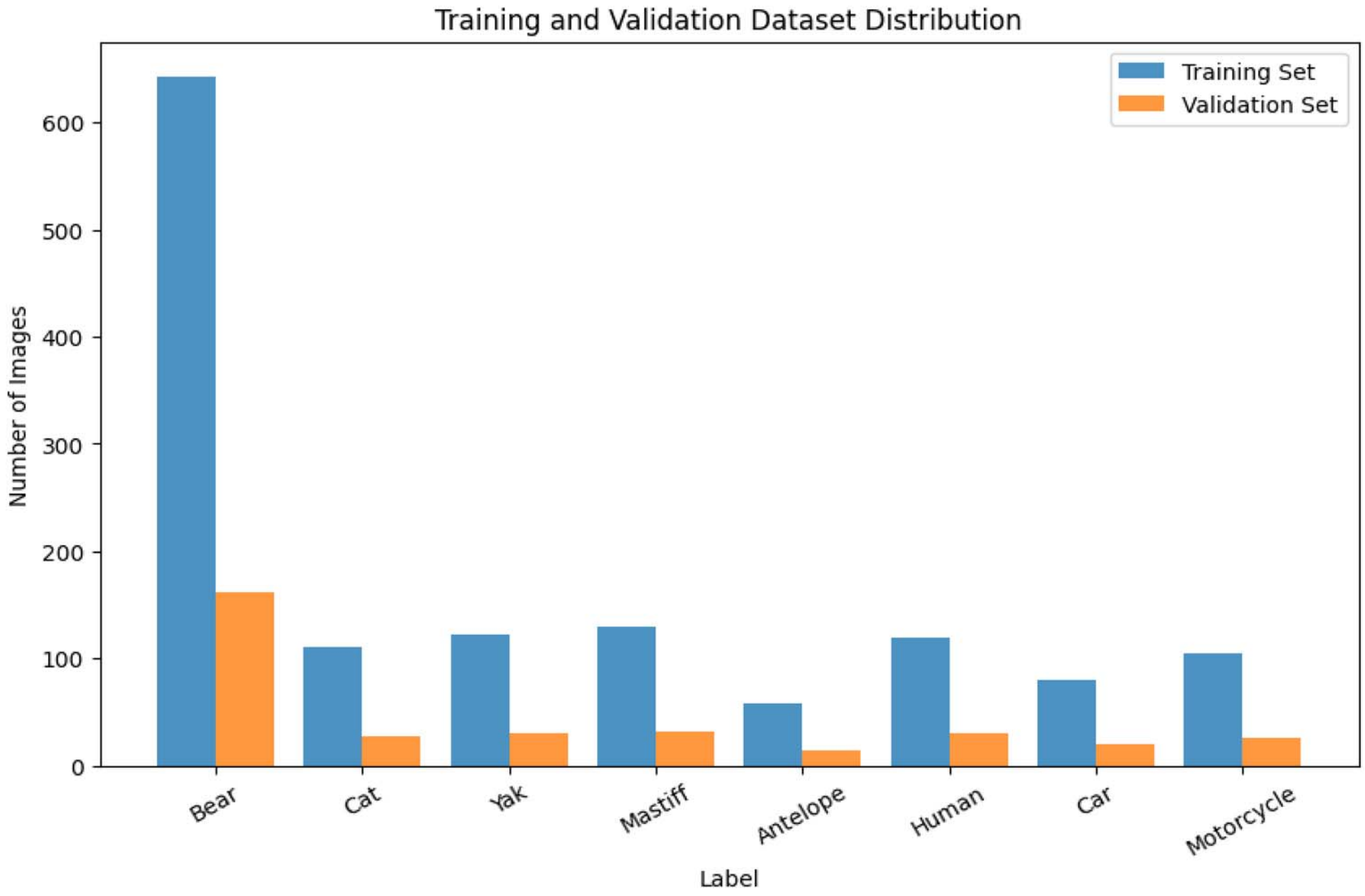

**Fig. 2. Number of images by subject type derived from a data set of 1,243 wildlife images representing species commonly encountered on the Tibetan Plateau, collected from field surveys and online sources, and used for training and validation of the bear detection model.**

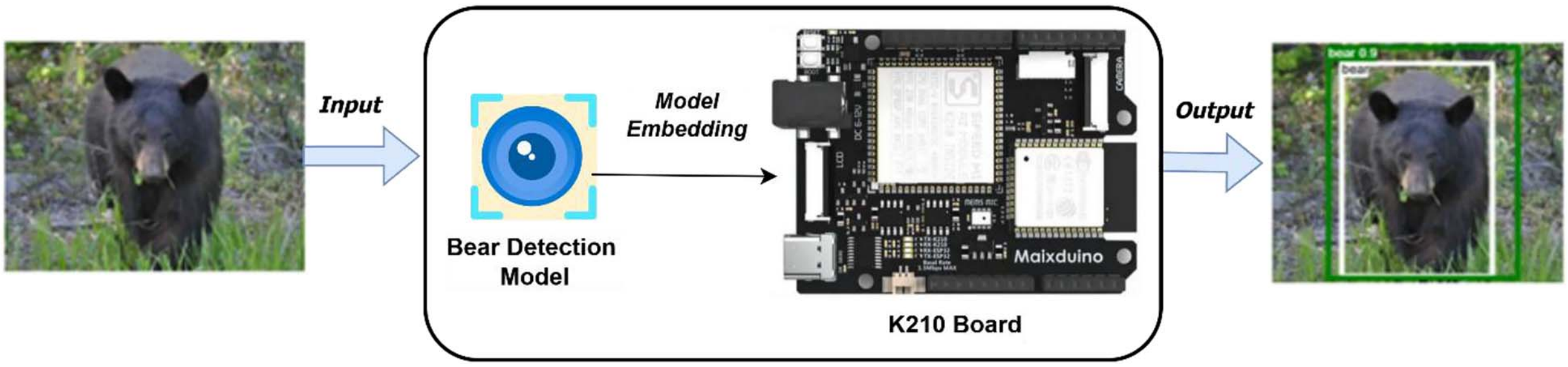

**Fig. 3. Workflow of the bear detection system that combines computer vision with Internet of Things (IoT) hardware. The trained model is deployed on a K210-based Maixduino board (Sipeed, Shenzhen, Guangdong, China), where incoming images are processed in real time. Detected bears are highlighted with bounding boxes and associated confidence scores.**

The trained model is embedded directly into the K210's onboard NPU. By processing images on-device, the system remains fully operational offline, which is critical for remote locations where network connectivity is limited or unavailable. The unit's typical power consumption ranges from 200 to 500 megawatts, supported by a 1-watt solar panel (18 × 14 cm) and an 11,000 milliampere-hour (mAh) lithium battery. During field testing, this configuration sustained continuous operation for 30 days, demonstrating reliable, long-term performance in off-grid environments. The system's average energy draw is lower than the panel's generation capacity, so it can operate approximately 4 days on battery alone and indefinitely under consistent solar recharging. The 30-day duration therefore reflects the length of the field experiment rather than a hardware limitation.

When the detection algorithm identified a bear, a specialized high-pressure bear spray device was activated for 1 second. The device operated at up to 2.8 megapascals (Mpa) and used a high-pressure nozzle capable of reaching approximately 13 m in still conditions, providing an extended protective range. In our tests, the detection module had a recognition range of about 10 m, which is shorter than the maximum spray distance. As a result, any bear detected by the system was already within the effective range of the spray, ensuring that deterrence could be reliably deployed. The spray itself includes 2–5% capsaicin and 1–2% menthol, which causes intense but short-lived discomfort. It is highly effective at stopping aggressive behavior and preventing injuries in close encounters with bears, thereby reducing conflict and the need for lethal measures. A negative encounter with bear spray may also train bears to avoid humans, potentially promoting more peaceful coexistence in shared habitats. The internal structure of the system, including the main components such as the K210 board, camera module, and bear spray mechanism, is shown in Figure 4.

**Table 1. Training configuration of the YOLOv5-MobileNet Bear Detection Model for a low-power, network-independent deterrence system that combines computer vision with Internet of Things (IoT) hardware. Model was derived from a data set of 1,243 wildlife images representing species commonly encountered on the Tibetan Plateau, collected from field surveys and online sources.**

| Image augmentation | Model information | Training parameters |
|---|---|---|
| Random mirroring: Enabled | Model type: Transfer learning | No. of training iterations: 100 |
| Random blur: Enabled | Model network: YOLOv5 | Batch size: 32 |
| Scaling method: Contain | Backbone network: MobileNet_0.75 | Learning rate: 0.001 |
| Scaling size: 224 × 224 | | Bounding box limit: 10 |

**Table 2. Hardware specifications of the On-Device Bear Detection Unit deployed on the Tibetan Plateau as part of a low-power, network-independent deterrence system that combines computer vision with Internet of Things (IoT) hardware.**

| Parameter | Specification |
|---|---|
| CPU & NPU[a] | RISC-V Dual Core 64-bit @ 400 MHz + FPU, AI NPU |
| Camera (GC0328) | QVGA@60 FPS, VGA@30 FPS, 90° × 60° FoV |
| Memory | 8-MB SRAM |
| Connectivity | 2.4-GHz Wi-Fi (802.11 b/g/n), Bluetooth 4.2 |
| Power consumption | ~200–500 mW |
| Solar & battery setup | 1-W solar panel, 11,000-mAh battery (≥30 days operation) |

[a]CPU is central processing unit; NPU is neural processing unit.

### *Testing and field implementation*

To assess the system's real-world performance, we conducted 100 activation tests across multiple outdoor environments in Zadoi County, Qinghai Province, China. We defined an activation test as a controlled approach toward the device by either a simulated target (person, mastiff, yak, goat, motorcycle) or by naturally occurring wildlife observed in the field. Each test session lasted from 5 to 10 minutes, depending on the number of approach scenarios performed. We conducted tests under both daytime and nighttime conditions to evaluate robustness across variable lighting environments.

For every trial, we recorded (1) whether the approaching subject was correctly detected, (2) whether the spray module activated, and (3) whether the activation outcome was appropriate (true positive deterrence, false positive, or false negative). This procedure allowed systematic evaluation of detection accuracy, latency, and mechanical reliability across different species and environmental conditions.

To further test long-term stability, 3 full systems were installed in locations with high bear activity based on local reports and community input: 1 unit near a landfill and 2 units near herders' homes (Fig. 5). Of the 100 activation tests, 40 were conducted at Site A (landfill area), 30 at Site B (near a herder's home), and 30 at Site C (riverbank site). All units operated autonomously, powered by solar panels and offline inference, ensuring consistent functionality in this remote, electricity-scarce region. The deployments lasted for 30 consecutive days (16 May–25 Jun 2023), overlapping with the activation trials. During this period, ambient temperatures ranged from −1°C to 18°C, with frequent rain events and marked diurnal variation. Across deployments, the systems maintained stable operation, and 3 successful bear deterrence events were recorded without false activations. The details of these events are described in the Results below.

**Table 3. Glossary of key technical terms related to the On-Device Bear Detection Unit deployed on the Tibetan Plateau during 2023 as part of a low-power, network-independent deterrence system that combines computer vision with Internet of Things (IoT) hardware.**

| Term | Description |
|---|---|
| Internet of Things (IoT) | A network of physical devices that integrate sensors, processors, and actuators to operate autonomously and exchange data without requiring continuous human control. |
| Computer Vision (CV) | A field of artificial intelligence that enables machines to interpret and analyze visual data (e.g., images, videos) for object detection and recognition. |
| Edge AI | Artificial intelligence computations performed locally on a device ("on the edge") rather than relying on cloud servers, reducing latency and ensuring offline functionality. |
| K210 Board | A low-power Kendryte K210 microprocessor designed for embedded AI applications, supporting real-time image recognition and device control in the field. |
| Activation test | A controlled trial in which a subject (e.g., bear, human, dog, livestock) approaches the system to test detection, spray triggering, and recorded outcome. |

## Results

### *Model training results*

The model exhibited high reliability for the validation data set, achieving a mean average precision (mAP) of 91.4% and a recall rate of 93.6%. mAP reflects the likelihood that a detected bear is indeed a bear (reducing false detection and alarms), while recall measures the model's ability to identify actual bear instances (ensuring most bear sightings are captured). Additionally, the model attained an F1 score (a machine learning evaluation metric that balances precision and recall to provide a comprehensive assessment of the model's performance) of 94.7%, indicating that only a small fraction of non-bear instances were misclassified as bears. During training, both training and validation accuracy curves showed stable convergence after approximately 100 epochs, confirming consistent learning and no evidence of overfitting.

Misidentification rates across non-bear classes were calculated from the confusion matrix of the validation data set (Table 4). Each rate reflects the percentage of non-bear instances erroneously identified as bears. The model occasionally confused morphologically similar animals, such as Tibetan mastiffs (6.4%) and yaks (5.1%), particularly under shadowed or semi-lit conditions. These results confirm that while nocturnal bear detection remained stable, future data set expansion should include additional nighttime samples of other species to further reduce misclassification risk.

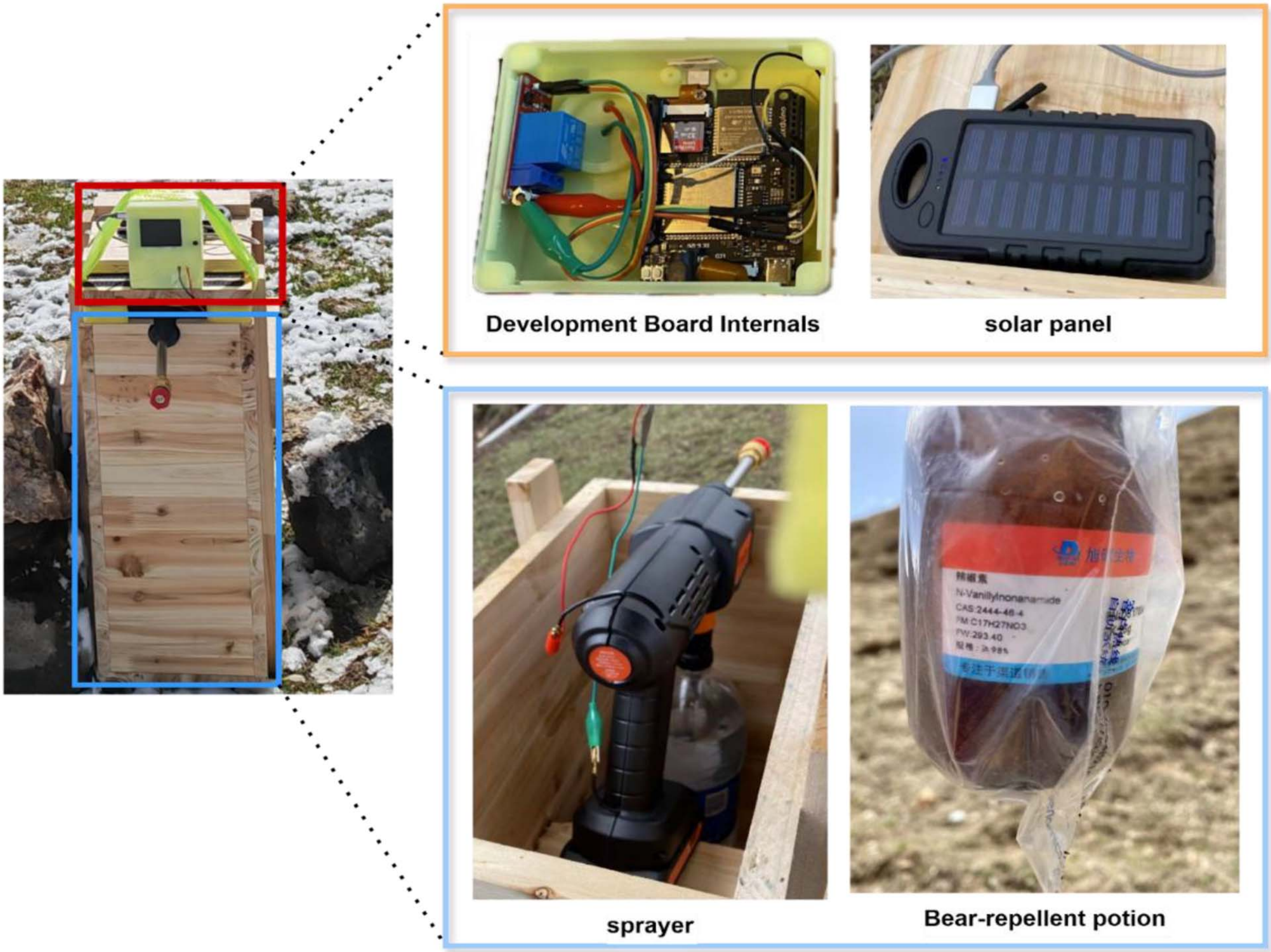

**Fig. 4. Key internal components of the bear deterrence system. The setup includes the K210 development board for on-device detection, a solar panel for off-grid power, a high-pressure sprayer, and a bear-repellent canister.**

### *Deterrence system performance*

The bear spray module integrated with the K210 board proved to be both reliable and efficient in controlled testing. The system demonstrated a high triggering accuracy of 97.2%, ensuring that the spray is activated only in response to a detected bear. The module operated with remarkable speed, deploying the spray within 0.2 seconds after detection, providing a timely deterrent to approaching bears.

We used a supplementary camera to monitor activity around the deployed systems. Over a 30-day testing period in the field, 3 successful Tibetan brown bear deterrence events were recorded by a monitoring camera integrated with a bear spray module. In all instances, the bears retreated immediately upon spray deployment. All events were recorded by the same unit, which had been strategically positioned near a water channel in a residential village, highlighting the camera's effectiveness in a high-risk location. Of these, 2 deterrence events occurred at night and 1 event during the day. Figure 6 captures a nighttime deployment, demonstrating that the detection model performs reliably under low-light conditions. No non-bear species were observed to trigger the device, and no false activations occurred during the deployment. These real-world observations reinforce the system's potential to serve

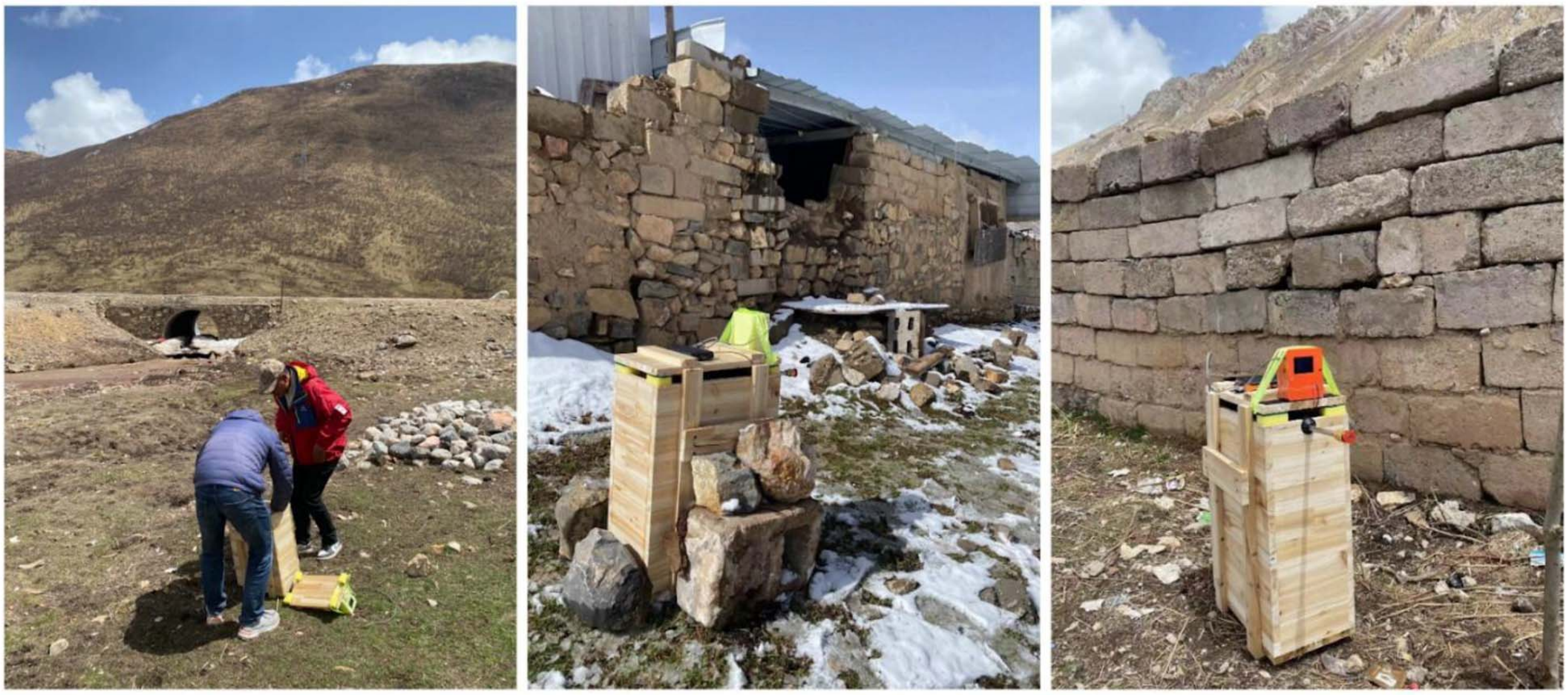

**Fig. 5. Deployment sites of the bear deterrence units in Zadoi County, Qinghai Province, China—1 unit placed near a local landfill and 2 placed near herders' residences during 2023. These sites were selected based on reported brown bear (*Ursus arctos pruinosus*) encounters and logistical feasibility for off-grid installation.**

as a proactive and nonlethal intervention tool in reducing human–wildlife conflict.

## Discussion

### *System contributions*

The proposed Intelligent Bear Prevention System addresses a critical need in the Tibetan Plateau region where human–bear interactions are increasingly common because of habitat overlap and seasonal migrations (Dai et al. 2019). By providing an automated, localized approach to detecting and deterring bears, the system can help to protect both local communities and wildlife.

**Table 4. Misidentification rates across non-bear categories in the validation data set derived from 1,243 wildlife images representing species commonly encountered on the Tibetan Plateau, collected from field surveys and online sources, and used for training and validation of the bear detection model.**

| Category | Misidentification probability (%) |
|---|---|
| Tibetan antelope (*Pantholops hodgsonii*) | 0.12 |
| Tibetan mastiff (*Canis familiaris*) | 6.40 |
| Cat (*Felis catus*) | 0.08 |
| Yak (*Bos mutus*) | 5.11 |
| Car | 2.07 |
| Motorcycle | 2.04 |
| Human | 1.76 |

Unlike traditional bear deterrence methods, which often rely on constant human presence or manual intervention, our solution offers a more sustainable and consistent form of conflict mitigation. Additionally, the data generated from the system (e.g., detection events, time logs) can inform wildlife management strategies, aiding researchers and local authorities in understanding bear movement patterns and behaviors.

### *Operational stability and sustainability*

The central strength of the system lies in its low-power design. The Kendryte K210-based edge AI board consumes minimal energy, and the unit operates autonomously on solar power. Field deployments demonstrated stable performance for $\geq$30 consecutive days (16 May–25 Jun 2023), despite temperature fluctuations from $-1$°C to 18°C and frequent rainfall. The current trial period was limited to 1 month, but the system is designed for indefinite operation provided that solar charging continues and spray canisters are replaced when exhausted. This off-grid independence makes the device particularly suited for remote, infrastructure-scarce areas.

### *Ethical, legal, and economic considerations*

Nonlethal deterrence is essential for wildlife conflict mitigation. The spray formulation uses safe concentrations

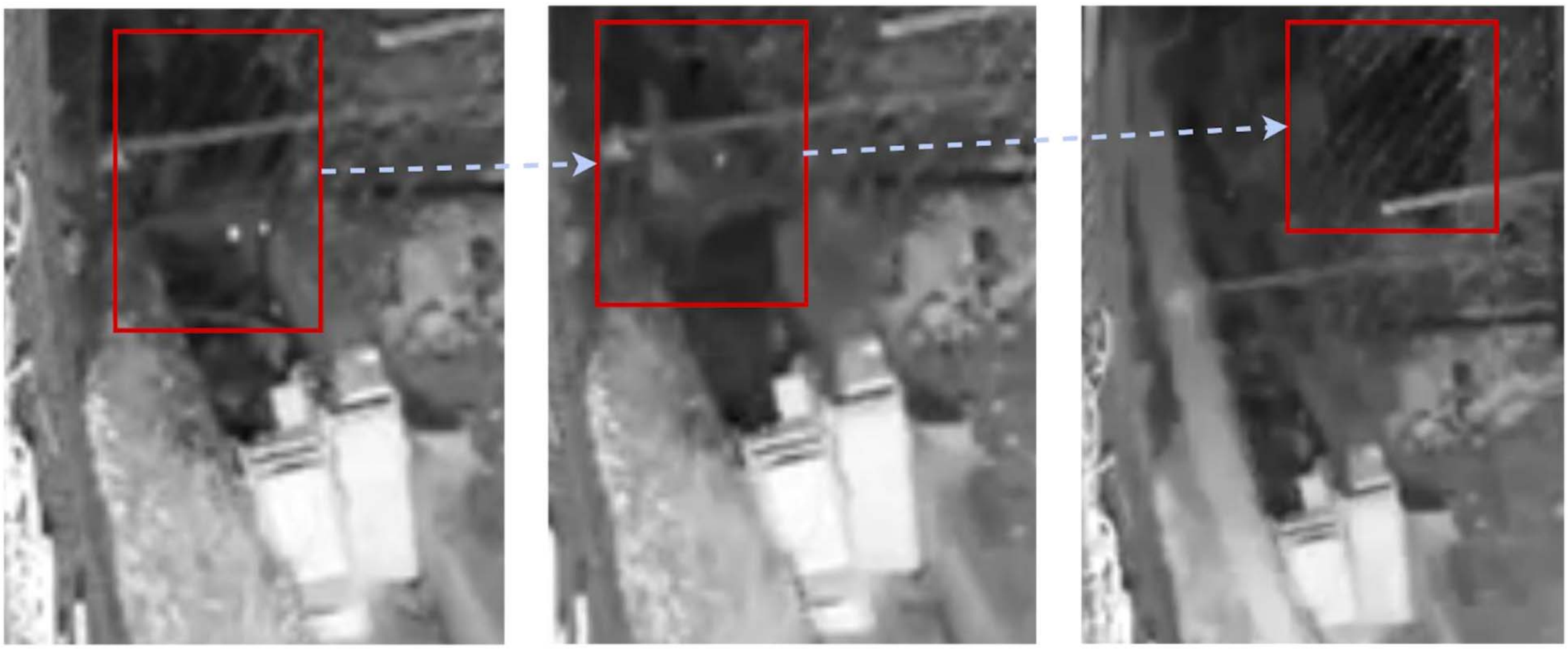

(a) The brown bear approaches the herder's residence. (b) The anti-bear system is activated. (c) The brown bear moves away from the residence.

**Fig. 6. A bear deterrence event captured in 2023 by a monitoring camera placed in Zadoi County, Qinghai Province, China. The brown bear (*Ursus arctos pruinosus*) is highlighted with a red bounding box.**

of capsaicin and menthol to repel bears while minimizing harm to unintended targets. This approach aligns with wildlife protection standards and local regulations intended to safeguard indigenous species. Furthermore, clear guidelines for system deployment (e.g., installation locations, signage, and community education) are necessary to ensure public safety, avoid accidental human exposure to the spray, and uphold local legal statutes regarding animal deterrents. Ongoing dialogue with conservation agencies and local governments remains essential to adapt the system ethically in diverse socio-ecological contexts (Dai et al. 2020b).

Cost is another factor supporting wide adoption: each unit is estimated at 470 Chinese yuan (RMB; 66 U.S. dollars [USD]; Table 5), making it a cost-effective option for widespread deployment in remote regions such as Qinghai Province. This affordable design significantly reduces financial barriers for adoption, especially for local herders and conservation agencies operating on limited budgets.

### Limitations

Several limitations must be acknowledged. Despite being diverse, the training data set remains modest in size, and expanding it across seasons and habitats would improve model robustness. In addition, it contains an imbalance in nocturnal imagery, and while >100 infrared bear images were included to ensure reliable nighttime detection, other species were primarily represented by daytime images. This imbalance may bias the model toward associating nighttime conditions with bears, slightly reducing its generalizability under diverse illumination settings. Future work will expand the data set to include additional nocturnal images of non-bear species to improve balance across lighting conditions.

**Table 5. Estimated cost breakdown of a bear detection and deterrence unit deployed on the Tibetan Plateau during 2023 as part of a low-power, network-independent deterrence system that combines computer vision with Internet of Things (IoT) hardware.**

| Component | Quantity | Unit price (RMB/USD)[a] | Total cost (RMB/USD)[b] |
|---|---|---|---|
| K210 development board | 1 | 200 / 28 | 200 / 28 |
| Camera | 1 | 50 / 7 | 50 / 7 |
| High-pressure spray nozzle | 1 | 60 / 8 | 60 / 8 |
| Solar panel | 1 | 50 / 7 | 50 / 7 |
| Electronic component kit | 1 | 30 / 4 | 30 / 4 |
| 11,000-mAh battery | 1 | 20 / 3 | 20 / 3 |
| Enclosure and mounting accessories | 1 | 20 / 3 | 20 / 3 |
| Bear spray canister | 1 | 30 / 4 | 30 / 4 |
| Cables and connectors | 1 | 10 / 2 | 10 / 2 |
| Total | 9 | | 470 RMB / 66 USD |

[a]RMB refers to Chinese yuan, and USD to United States dollar.
[b]Quoted costs are from Apr 2025.

Only 3 confirmed deterrence events were recorded during field trials, limiting the scope of evaluation for long-term effectiveness. Spray dispersal may be influenced by wind and terrain, factors requiring further investigation. Additionally, long-term durability across multiple years and extreme conditions remains untested.

Another important limitation relates to the choice of detection model. We deployed YOLOv5 because it remains one of the most efficient architectures available for small, low-power IoT devices. Although more recent versions such as YOLOv12 achieve higher accuracy in research benchmarks, they require significantly greater computational resources. Deploying these models in our current hardware setup would demand upgrades in both processing power and energy supply, potentially exceeding the capabilities of the solar-based system. Balancing accuracy with energy efficiency remains an ongoing challenge for real-world, off-grid applications.

### *Future improvements and broader applicability*

Future iterations of the system could integrate more advanced IoT models and hardware to further enhance performance. Incorporating stronger AI chips would enable faster real-time processing, while mechanical arms could dynamically adjust spray direction, reposition sensors, or deploy alternative deterrents such as sound alarms or flashing lights. Detection reliability could also be improved by integrating additional modalities such as thermal or acoustic sensors, particularly under low-visibility conditions. Extending deployments across multiple sites and seasons would generate long-term data to refine both detection algorithms and deterrence strategies.

The system's affordability, autonomy, and modular design suggest broad potential for application beyond the Tibetan Plateau. Similar frameworks could be adapted for human–wildlife conflict involving grizzly bears (*U. arctos horribilis*), European brown bears (*U. a. arctos*), or other large carnivores in rural regions. By partnering with conservation organizations and local communities, the technology could be tailored to diverse ecological and cultural contexts, scaling its impact while balancing human safety and wildlife conservation.

## Conclusion

This study presents a computer-vision-based, low-power system that integrates IoT technologies to address human–bear conflicts in remote, off-grid environments. Leveraging lightweight models and solar-powered operation, the system provides continuous monitoring and automatic, nonlethal deterrence, ensuring the protection of herders' property and safety. From a wildlife management perspective, the system provides a sustainable and humane alternative to traditional methods such as mesh wire fences or hunting. By avoiding the ethical and legal risks associated with harming protected species, it ensures compliance with conservation regulations while effectively safeguarding local herders' properties. By deterring bears from human settlements and artificial food sources, the system helps reduce bears' reliance on these resources (Dai et al. 2020b). This can positively influence their long-term behavior, encouraging a return to more natural foraging habits and contributing to healthier wildlife populations, while promoting a balanced coexistence between humans and bears.

## Acknowledgments

This research was supported by the Qinghai Provincial Department of Science and Technology, the Institute of Zoology at the Chinese Academy of Sciences, and the Key Program of the Open Fund of the State Key Laboratory of Resources and Environmental Information Systems (2023). We thank the local communities in Zadoi County for their assistance during fieldwork and deployment. In particular, we thank Mr. D. Sangji for his invaluable assistance in introducing us to local stakeholders and providing transportation to all test sites. We also acknowledge the use of OpenAI's GPT-4o model for language. We thank the Associate Editor and reviewers for their suggestions. Preliminary findings from this study were previously made available as a preprint (Chen et al. 2025).

## Author contributions

P. Chen: Conceptualization, system design, field deployment, writing—original draft; F. Teng: Conceptual guidance, supervision, writing—review & editing; J.A. Kupfer: Wildlife expertise, supervision, writing—review & editing; Y. Du, J. Yi, Y. Li: Field deployment, local coordination. All authors reviewed and approved the final version of the manuscript.

## Ethics statement

No people and animals were harmed during this research. All deterrent measures employed were nonlethal and formulated to minimize stress to wildlife. The bear spray composition complied with national safety standards. Field activities were conducted with consent from residents and in accordance with ethical guidelines for wildlife conservation and human–wildlife conflict mitigation.

It is important to note that the system does not store or transmit images. All processing occurs locally on the edge device, and only bear-related detection events are logged as metadata without visual records. Consequently, human images are not retained, minimizing concerns regarding personal data protection. The supplementary monitoring camera also did not record any human presence during the field trial. Future developments could integrate additional privacy safeguards, such as automated filtering of human images if they were ever captured, to ensure compliance with international standards on data protection.